\documentclass[a4paper, fleqn]{cas-dc}
\usepackage[inkscapelatex=false]{svg}
\usepackage{svg} 
\usepackage{float}
\usepackage{graphicx}
\usepackage{subcaption}
\restylefloat{figure}

\usepackage[numbers]{natbib}

\def\tsc#1{\csdef{#1}{\textsc{\lowercase{#1}}\xspace}}
\tsc{WGM}
\tsc{QE}

\begin{document}
\let\WriteBookmarks\relax
\def\floatpagepagefraction{1}
\def\textpagefraction{.001}
\let\printorcid\relax 

\shorttitle{RETHINKING SUPERPIXEL SEGMENTATION FROM BIOLOGICALLY INSPIRED MECHANISMS}    

\shortauthors{TingYu zhao et al.}

\title[mode = title]{RETHINKING SUPERPIXEL SEGMENTATION FROM BIOLOGICALLY INSPIRED MECHANISMS}  

\tnotemark[1,2]

\tnotetext[1]{This work was supported by Natural Science Foundation of Sichuan China (No. 2022NSFSC0502), National Science Foundation of China (No. 42075142) and Key Research and Development Program of Sichuan Province(No. 2023YFG0125).}

\author[1]{TingYu Zhao}
\fnmark[1]
\author[1]{Bo Peng}
\cormark[1] 
\fnmark[2]
\author[2]{Yuan Sun}
\fnmark[3]
\author[1]{DaiPeng Yang}
\fnmark[4]
\author[1]{ZhenGuang Zhang}
\fnmark[5]
\author[3]{Xi Wu}
\fnmark[6]

\address[1]{School of Computing and Artificial Intelligence, Southwest Jiaotong University, Chengdu 610031, China}
\address[2]{School of Physics, Nankai University, TianJing 300071, China}
\address[3]{School of Computer Science, Chengdu University of Information Technology, Chengdu 610031,
China}

\cortext[1]{Corresponding author} 

\begin{abstract}
Recently, advancements in deep learning-based superpixel segmentation methods have brought about improvements in both the efficiency and the performance of segmentation. However, a significant challenge remains in generating superpixels that strictly adhere to object boundaries while conveying rich visual significance, especially when cross-surface color correlations may interfere with objects. Drawing inspiration from neural structure and visual mechanisms, we propose a biological network architecture comprising an Enhanced Screening Module (ESM) and a novel Boundary-Aware Label (BAL) for superpixel segmentation. The ESM enhances semantic information by simulating the interactive projection mechanisms of the visual cortex. Additionally, the BAL emulates the spatial frequency characteristics of visual cortical cells to facilitate the generation of superpixels with strong boundary adherence. We demonstrate the effectiveness of our approach through evaluations on both the BSDS500 dataset and the NYUv2 dataset.

\end{abstract}




\maketitle

\section{Introduction}

Superpixel segmentation algorithms transform the original image into distinct regions with visual significance. During the transition from pixel-level to region-level representation, the process not only clusters similar pixels at a low level, but the generated superpixels also retain independent semantic information. As a result, such algorithms improve the efficiency of image representation and reduce data dimensionality, making them indispensable tools for various computer vision tasks, including semantic segmentation\cite{he2015supercnn} \cite{yang2013saliency} \cite{zhu2014saliency}, object detection\cite{sharma2014recursive}\cite{gadde2016superpixel}, optical flow estimation\cite{lu2013patch} \cite{hu2016highly} \cite{sun2014local} \cite{yamaguchi2013robust}, among others.

Pixel label assignment in superpixel segmentation is determined by their associations with surrounding neighborhoods. Obtaining accurate relational associations is paramount in superpixel segmentation. Traditional methods of superpixel segmentation often utilize clustering or graph-based techniques \cite{achanta2012slic}\cite{achanta2017superpixels}\cite{li2015superpixel}\cite{liu2011entropy}\cite{liu2016manifold} to determine pixel affiliations with adjacent grids. However, these handcrafted feature-dependent methods struggle to achieve optimal boundary adherence and to maintain semantic information for superpixel segmentation. With the remarkable success of deep learning, several deep neural network (DNN)-based superpixel segmentation algorithms have emerged. For example, the SEAL\cite{tu2018learning} algorithms extract high-level abstract features through artificial neural networks, replacing traditional handcrafted features like color, depth and shape attributes. Additionally, SSN\cite{jampani2018superpixel} introduced a soft relationship between pixels and grids to replace the nearest-neighbor algorithm, thereby rendering the SLIC\cite{achanta2012slic} algorithm differentiable and addressing the inefficiencies linked to standard convolution on irregular superpixel grids. By leveraging a fully convolutional network, SCN\cite{yang2020superpixel} predicts the association scores between image pixels and grids end-to-end without clustering. This innovation enhances the efficiency of superpixel segmentation algorithms further. Recently, algorithms like AINet\cite{wang2021ainet} correlate high-level grid features with corresponding pixels, supplying potent contextual information. Moreover, OversNet conducts a task-driven neural network search to identify the most fitting convolutional kernels for the encoder.

However, these DNN-based algorithms struggle to learn accurate association maps utilizing limited training data, due to factors such as cross-obstruction, noise, and artifacts in the images. The generated superpixels under such conditions can be blunt, i.e., including pixels of different semantics. In contrast, the human eye is highly sensitive to subtle disturbances and can swiftly discern object boundaries and hierarchical relationships in natural images with complex textures. Moreover, it has been evidenced that artificial neural networks can adeptly emulate the biological visual system, relevant particularly to the ventral visual stream in primates\cite{bashivan2019neural} \cite{schrimpf2018brain}.

To address the aforementioned challenges, we draw inspiration from the neural structure and biological visual mechanisms to propose a novel biomimetic network for superpixel segmentation. This network endeavors to align the learned features with human-recognizable characteristics as closely as possible. Specifically, guided by the multilevel synchronization hypothesis of the visual cortex\cite{zeki1993vision}, we propose the Enhanced Screening Module(ESM), designed to emulate human visual information processing. This module fuses low-level texture and color details with high-level semantic data in the decoder network module, facilitating interactive, multilevel complex visual information processing and capturing more precise and robust relationships between pixel grids. Furthermore, based on the band-pass frequency sensitivity characteristics of the visual cortex cell \cite{blakemore1969existence}—highlighting the significance of edge information due to high spatial change rates and low contrast thresholds at edges—we propose the innovative Boundary-Aware Label (BAL). The BAL amplifies the network's responsiveness to boundary, leading to the generation of more precise superpixels that align closely with semantic boundaries.

The main contributions are summarized as follows: 
\begin{itemize}

\item The Enhanced Screening Module is proposed to integrate features such as color, depth and shape in a multilevel synchronization framework, facilitating the generation of strongly semantically correlated superpixels.

\item The Boundary-Aware Label is proposed to  enable the network to pay more attention to boundary disturbances and make superpixels have good boundary adherence.

\item We demonstrate that two mechanisms of the human visual system consistently enhance the accuracy of superpixel segmentation. Extensive experiments on the open databases are conducted to verify the performance of the proposed network.





\end{itemize}

\section{Related Work}
\label{sec:related}
\subsection{Superpixel Segmentation}
 Superpixels segmentation have undergone substantial advancement since superpixels were first proposed in \cite{ren2003learning}. Traditional superpixel segmentation is categorised into graph-based and gradient descent methodologies. Notable graph-based techniques encompass EGBIS\cite{felzenszwalb2004efficient}, Ncuts\cite{shi2000normalized}, and the entropy rate superpixels(ERS)\cite{liu2011entropy}. Within this paradigm, pixels are conceptualised as nodes, inter-pixel relationships as edges, and feature similarities as edge weights. Employing the principle of a minimum spanning tree, images are hierarchically segmented. However, these strategies either face challenges in generating uniform superpixels or adequately preserving image boundaries; Gradient descent based methods use clustering ideas to generate superpixels. Similar to this paper, methods based on regular grid initialization mainly include the Simple Linear Iterative Clustering(SLIC)\cite{achanta2012slic} and Super-pixels Extracted via Energy-Driven Sampling(SEEDS)\cite{van2012seeds}. Variants of SLIC include Linear Spectral Clustering(LSC)\cite{li2015superpixel}, Manifold SLIC(MSLIC)\cite{liu2016manifold}and SNIC\cite{achanta2017superpixels}.These methods update cluster centers and pixel labels iteratively using gradient descent. After the rise of deep learning techniques, SEAL\cite{tu2018learning} adopted convolutional neural networks to learn deep features to cluster pixels, which is still a non-differentiable superpixel segmentation method. Recently proposed methods such as SSN\cite{jampani2018superpixel}, SCN\cite{yang2020superpixel} and AINet\cite{wang2021ainet} are the end-to-end trainable network for superpixel segmentation. These methods further improve the inference efficiency and performance of superpixel segmentation for downstream tasks. However, artificial neural networks can struggle to exhibit strong learning capabilities and robustness with limited data. As a contrast, we explore two biological mechanisms as the promising candidates to improve network learning ability.  Further, none of these works has attempted to jointly learn superpixels with the biological vision mechanisms.

\subsection{Application of Superpixel}

Superpixel segmentation permits the substitution of the original pixel-level annotations with fewer pixel blocks in the deep learning pipeline, diminishing storage requirements and annotation overheads while preserving important image properties (e.g., boundaries) \cite{gaur2019superpixel}\cite{sun2018spsim}\cite{yeo2017superpixel}\cite{li2012segmentation}\cite{bodis2015superpixel}. For example, Kim et al.\cite{kim2023adaptive}proposed an active learning method based on superpixels for semantic segmentation tasks. Zhu et al.\cite{zhu2023ecrf} integrate superpixels into E-CRF serving as an auxiliary to superpixel is integrated into E-CRF and served as an auxiliary for more reliable message passing between pixels in high-level features. The Superpixel-mix method proposed by Franchiet et al.\cite{franchi2021robust} incorporated superpixels into images as a mixed augmentation within teacher-student consistency training, aiming to enhance the awareness of object boundaries in semi-supervised semantic segmentation networks.  Liu et al.\cite{liu2018deep}
transformed superpixel segmentation outcomes into pixel-level results by computing the feature similarity between adjacent pixel blocks, which facilitates pixel-level tasks like semantic segmentation and edge detection.

\begin{figure}
	\includegraphics[width=0.85\linewidth,scale=1.5]{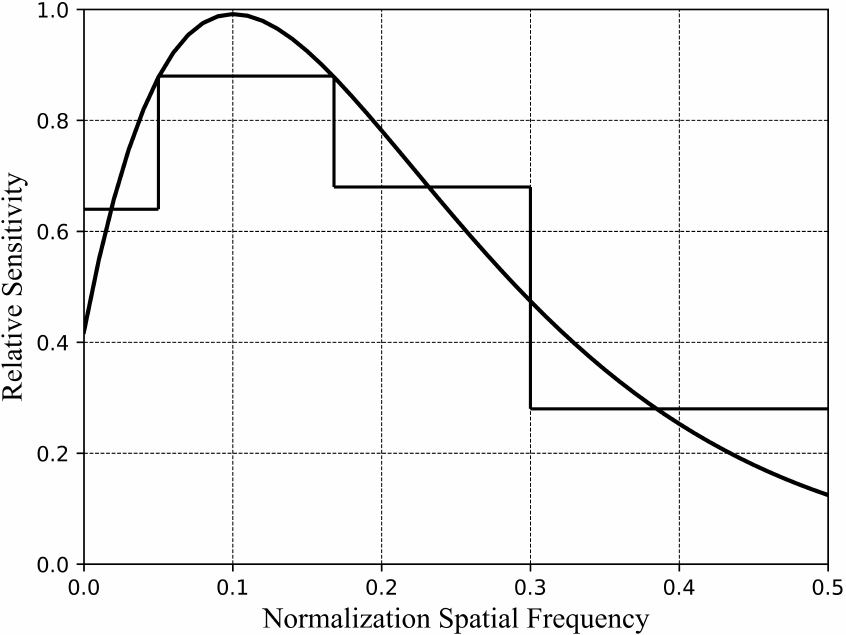}
	\centering
	\caption{The CSF Masking.}
	\label{fig:csf}
\end{figure}

\begin{figure}[t]
	\includegraphics[width=0.95\linewidth,scale=1.5]{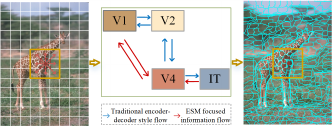}
	\centering
	\caption{Superpixel segmentation in the human visual system. \textbf{From left to right:} Input, the ventral visual pathway, Output. The ESM simulates information processing in the ventral visual pathway to more accurately predict the relationship between a pixel and its nine neighbor pixel grids.}
	\label{fig:ventral}
\end{figure}

\begin{figure*}
	\includegraphics[width=0.95\linewidth,scale=1]{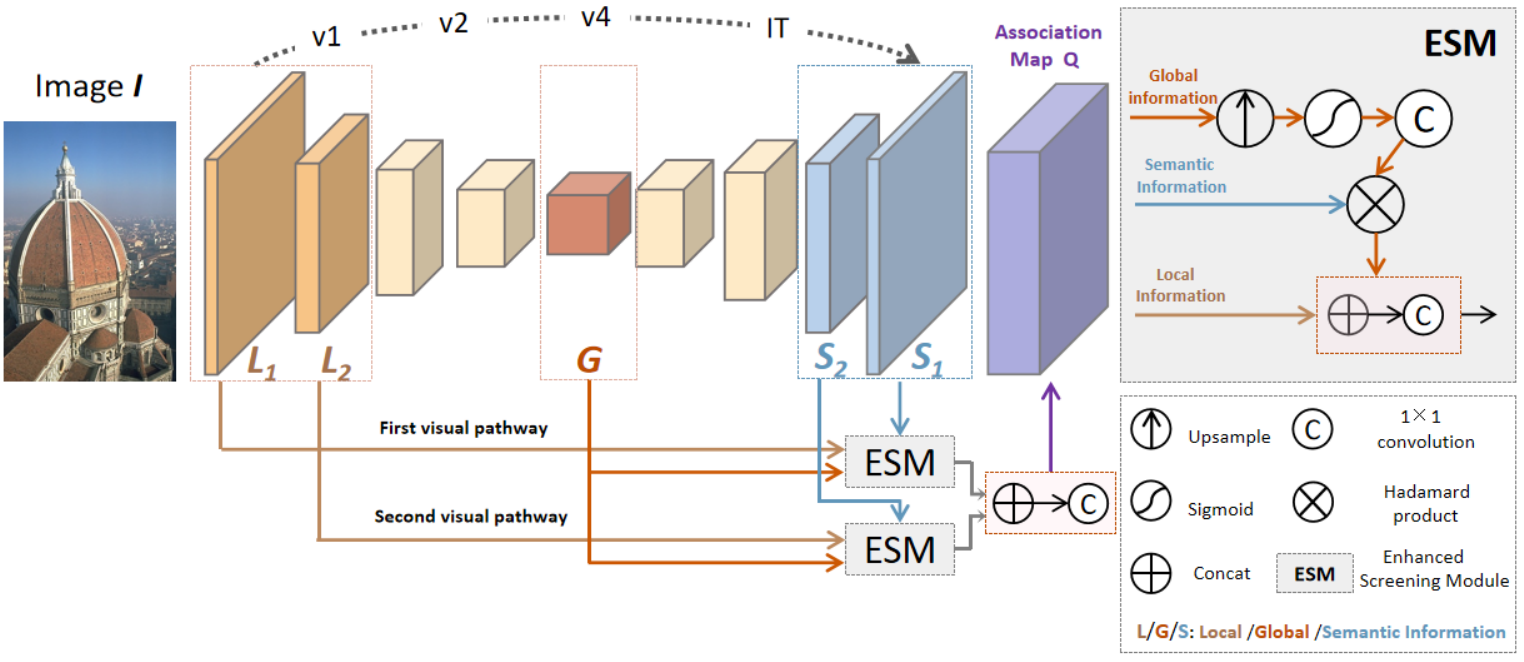}
	\centering
	\caption{The overall pipeline of the proposed biomimetic network architecture that contains two cascaded ESM modules, which correspond to two visual pathways capturing low-level features at distinct levels into the IT region, aiding in establishing enhanced contextual relationships.}
	\label{fig:network framework}
\end{figure*}	

\section{THE PROPOSED METHODS}
\label{sec:methods}
\subsection{Biologically Inspired Mechanisms}
Superpixel segmentation aims to obtain pixel blocks that maintain semantic information integrity as much as possible. For instance,  When segmenting different fine-grained(number of superpixels) images, it is essential to establish hierarchical relationships from person to face to eye and so on. Existing superpixel segmentation methods based on encoder-decoder architectures struggle to simultaneously capture low-level  and high-level features contextual relationships. Nevertheless, the human visual system demonstrates the ability to perceive small disturbances of color and texture. In this paper, we incorporate two biologically inspired visual mechanisms into the superpixel segmentation network to improve the accuracy of association maps. We present these two inspired visual mechanisms separately in this section. 

The first mechanism involves the multilevel synchrony integration hypothesis proposed by Zeki et al. \cite{zeki1993vision}, which suggests that the visual cortex integrates information by enlarging visual receptive fields to gather information across the entire visual field. As illustrated in Figure \ref{fig:ventral}, the integration mechanism exists in the ventral visual pathway of the visual cortex, i.e. through V1, V2, and V4 to reach the inferior temporal (IT) cortex. Specifically, The visual cortex progressively develops more complex and specialized properties, unifying information from different visual functions across various visual cortical areas. The human visual system outperforms at handling the associations between pixels and their surrounding pixel grids, encompassing both low-level textures and high-level semantic information. Additionally, studies on visual neurophysiology have indicated the presence of interactive feedback mechanisms between different organizational structures within the ventral stream \cite{nurminen2018top}\cite{choi2018bottom}. 

Hence,  we adapt their implementation to work with multiple-level information integration. Specifically, we map the superpixel segmentation network to the ventral visual pathway,
and the encoder-decoder network structure only contains part of the visual pathway (indicated by the blue line segments). We propose ESM(corresponding to the red line segments), which allows the network to capture the correspondence between low-level and high-level features by adding information processing pathways, instead of only focusing on some distinguishable features. Finally, it helps the model accurately predict the association between pixels and their nine adjacent superpixels.

The second mechanism relates to the frequency sensitivity and bandpass characteristics observed in the human visual system during image observation. Spatial frequency analysis methods have been widely employed in visual research, revealing that individual simple cells are sensitive to a specific range of spatial frequencies. This suggests the existence of neural information channels in the human visual system that analyze various spatial frequencies \cite{blakemore1969existence}. The visual contrast sensitivity function (CSF) is shaped by integrating envelopes from independent frequency-selective channels. On this foundation, Mannos and Sakfision proposed a classic model for the CSF\cite{poggio1985iii}\cite{petrovic2007objectively}, as follows:
\begin{equation}
\centering
H(f)=2.6 *(0.192+0.114 f) * e^{\left[-(0.114 f)^{1.1}\right]}.
\label{eq_csf}
\end{equation}
In Eq.(\ref{eq_csf}), the normalized spatial frequency is represented as $f=\sqrt{f_x^2+f_y^2}$ (cycles per degree), where $f_x$ and $f_y$ denote the spatial frequencies in the horizontal and vertical directions, respectively. As illustrated in Figure \ref{fig:csf}, the CSF curve reveals that human sensitivity exhibits non-linear variations with spatial frequency. Notably, the visual system displays pronounced sensitivity to edges due to their pronounced spatial variations and reduced contrast thresholds. Thus, the human eye is acutely attuned to edge details. Furthermore, artificial neural networks have been proven efficacious in simulating the biological visual system, showing relevance to the ventral visual pathway in primate brains. In sections \ref{ESM} and \ref{BAL}, we will elucidate the proposed methodologies, considering these two biologically inspired mechanisms. 

\subsection{Enhanced Screening Module}
\label{ESM}
An encoder-decoder framework, akin to AINet\cite{wang2021ainet} and SCN\cite{yang2020superpixel}, is employed to predict the association map in an end-to-end manner, i.e, the 9-way probabilities. Similarly, the centers of superpixels are ascertained via an element-wise multiplication between the predicted correlation map and the supplied feature map. Specifically, drawing inspiration from the hypothesis of multi-level integration mechanisms observed in human visual cortices, the Enhanced Screening Module (ESM) is proposed to support the decoder module. This facilitates the assimilation of salient cues from both local and global perspectives, enriching the construction of association relationship between adjacent grids.

As shown in Figure \ref{fig:network framework}, the entire biomimetic network architecture can be viewed as the propagation of visual signals through the ventral pathway. The encoder module is similar to the analysis and extraction of signals observed in retinal regions \emph{v1}, \emph{v2}, and \emph{v4}, whereas the decoder module resembles the perception of abstract shapes and colors in the Inferior Temporal(IT) area. 
\textbf{L}($L_{1}, L_{2}$) and \textbf{G} in the Figure \ref{fig:network framework} represent the local and the global information, respectively, while \textbf{S}($S_{1}, S_{2}$) denotes abstract semantic information. Notably, in \cite{wang2021ainet}, the AI module combines \textbf{G} and \textbf{$S_{1}$} via addition, which can be seen as a special variant of ESM. The biomimetic network architecture incorporates two overlaid ESM, effectively introducing two novel visual pathways that emulate the interactive projection mechanisms in the human visual cortex.

Formally, given an input image $\mathcal{I} \in \mathcal{R}^{H \times W \times 3}$, the local information ($L_{1} \in \mathcal{R}^{H \times W \times 16}$ and $L_{2} \in \mathcal{R}^{\frac{H}{2} \times \frac{W}{2} \times 32}$) represent primary color, texture and shape information in the visual cortex. With the elevation of cortical levels (indicated by the gray dashed arrow), each cortical cell correlates with more advanced levels of visual analysis. In the ventral information pathway, the global information ($G \in \mathcal{R}^{\frac{H}{16} \times \frac{W}{16} \times 128}$) symbolizes cellular responses to higher-level complex features such as facial and palmar patterns. The IT region, receiving visual data from several visual cortices, plays a pivotal role in shape perception and object recognition, corresponding to the decoder module in network architectures. Indeed, numerous lateral and feedback inputs establish intersecting pathways, enabling IT  region to access a relatively comprehensive set of visual information of varying complexity. Consequently, the information enhancement module combines the $L_{1}$, $L_{2}$, and $G$ information to influence the composition of the last two layers ($S_{1} \in \mathcal{R}^{H \times W \times 16}$ and $S_{2} \in \mathcal{R}^{\frac{H}{2} \times \frac{W}{2} \times 32}$) of the association map $Q$, facilitating cross-pathway information flow.  Let $\mathcal{C}(\cdot)$ and $\mathcal{D}(\cdot)$ denote the convolution operation and the deconvolution operation respectively. The two ESMs ($M_{1}$ and $M_{2}$)process can be represented as follows.
\begin{align}
M_{1}=[\mathcal{C}(\mathcal{D}(G))\otimes S_{1} ]\oplus L_{1}, \\
M_{2}=[\mathcal{C}(\mathcal{D}(G))\otimes S_{2} ]\oplus L_{2}.
\end{align}

The $M_{1}$ and $M_{2}$ scales are aggregated through a concatenation operation to generate the map $Q$. Compared to the approach using stacked convolutional blocks end-to-end, the proposed architecture to captured pixel correlation information in a biomimetic manner.

\begin{figure}[t]
	\includegraphics[width=0.95\linewidth,scale=1.5]{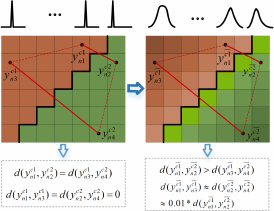}
	\centering
	\caption{A visual comparison between BAL and one-hot label.($C1$ and $C2$ denote two categories, $n1$, $n2$, $n3$, $n4$ denote pixel indexes respectively, with one-hot label on the left and the proposed BAL on the right.)}
	\label{fig:bal_com}
\end{figure}

\begin{figure}[t]
	\centering	
	\renewcommand{\tabcolsep}{4pt}
	\begin{tabular}{@{\hspace{-0.30cm}}c@{\hspace{0.05cm}}c@{\hspace{0.05cm}}c}		
		\includegraphics[width=0.33\linewidth]{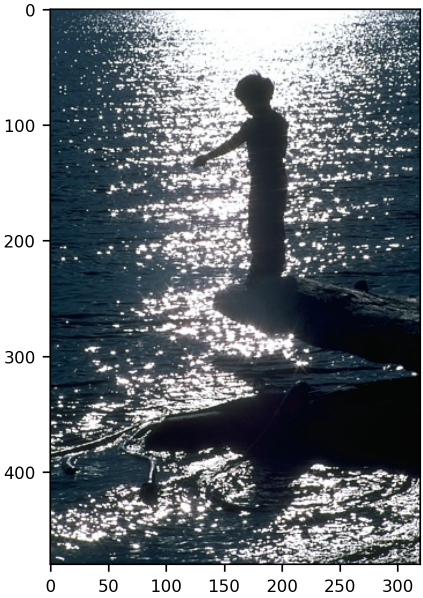}&		
  		\includegraphics[width=0.33\linewidth]{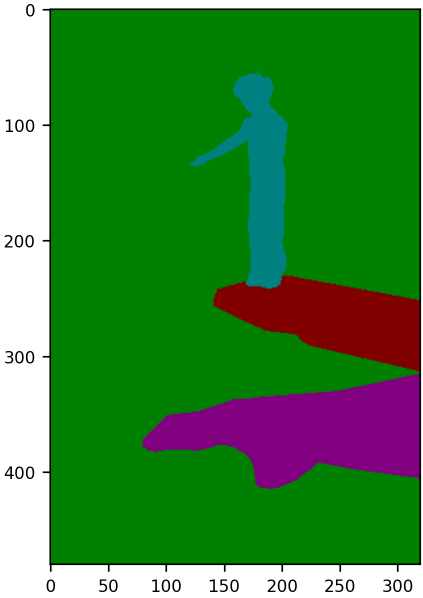}&		
		\includegraphics[width=0.39\linewidth]{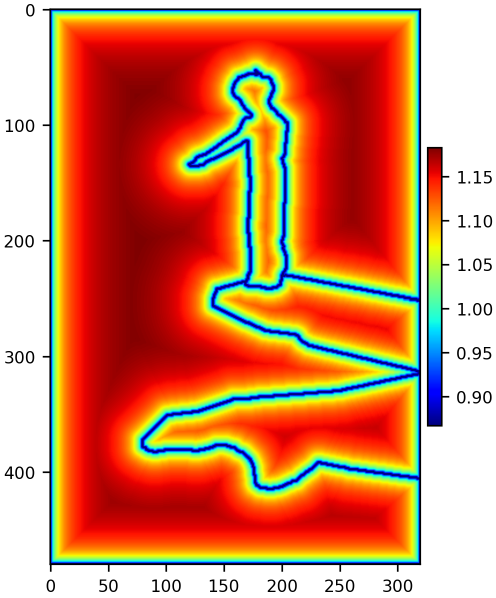}\\				
	    (a)  &(b) &(c) 
	\end{tabular}
	\caption{Visualization of the relationship between variance and distance. From left to right: (a) Image, (b) Ground Truth, and (c) Heatmap of Variance.}
	\label{sima2boundary}
\end{figure}

\begin{figure}[t]
	\centering	
	\renewcommand{\tabcolsep}{4pt}
	\begin{tabular}{@{\hspace{-0.05cm}}c@{\hspace{0.05cm}}c}		
		\includegraphics[width=0.80\linewidth]{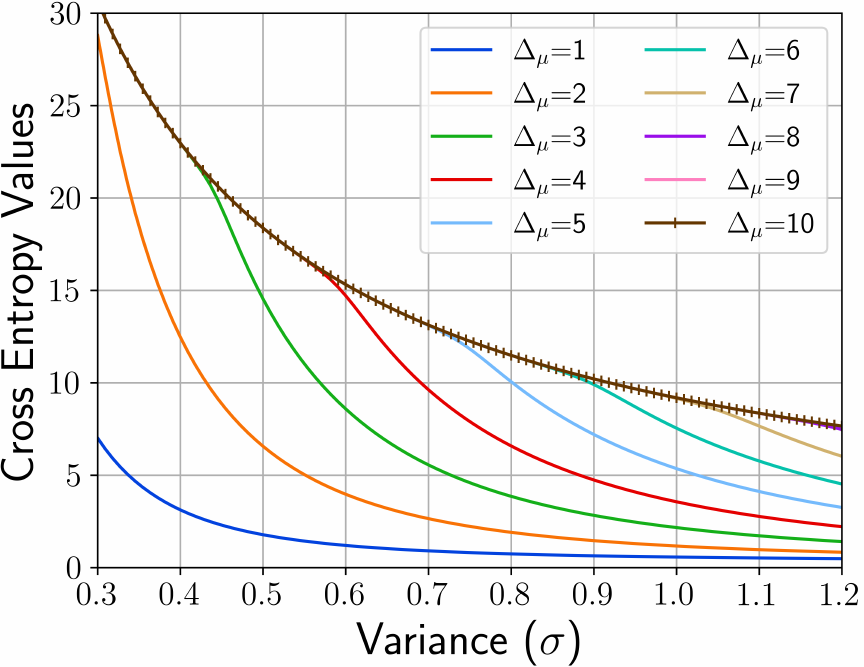}\\
            (a) \label{sigma} \\
  		\includegraphics[width=0.80\linewidth]{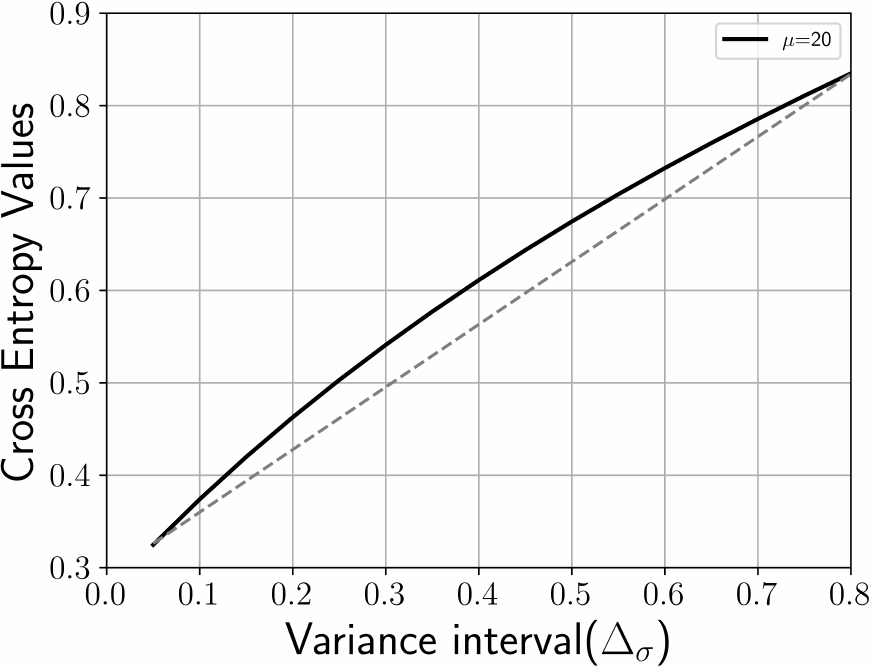}\\	
	    (b) \label{mean}
	\end{tabular}
	\caption{Quantitative Analysis of Cross-Entropy Distances in Gaussian Distributions.($\Delta_\mu$ and $\Delta_\sigma$ denote the subtraction of the mean and the subtraction of the variance respectively.)
(a) Illustrates variations in distances across $\Delta_\mu$ ranging from 1 to 10 and $\Delta_\sigma$ spanning from 0.3 to 1.2. Notably, distances remain constant when the $\Delta_\mu$ exceeds 10.
(b) Depicts alterations in loss while maintaining a constant mean value ($\mu$=20) and varying the $\Delta_\sigma$ from 0.05 to 0.8. }
\label{sigmaandmean}
\end{figure}

\subsection{Boundary-Aware Label} 
\label{BAL}
Based on the CSF model, we observe that the contrast sensitivity of the human eye changes with spatial frequency, demonstrating a bandpass characteristic. As the spatial frequency of image grayscale increases, its contrast threshold diminishes. Simultaneously, the noise's visibility decreases monotonically with spatial frequency. Consequently, the human eye is particularly attuned to spatial changes at edges. In this paper, leveraging the visual threshold effects inherent to the human eye, we propose a novel Boundary-Aware Label (BAL) to generate superpixels with enhanced boundary adherence. 

The proposed BAL is intended to endow the model with distinct sensitivities at edges versus within object interiors. Meanwhile, the label encoding strategy must also conform to specific constraints, specifically, ensuring that the inter-class distance between pixels at the boundary exceeds that between pixels farther from the boundary, while maintaining minimal perturbation in intra-class distances to preserve semantic information integrity. The gaussian distribution is employed for encoding label in order to enhance the robustness of the model against accross-surface color correlation disturbances.  As depicted in Figure \ref{fig:bal_com}, We visually demonstrate the changes in intra-class and inter-class pixel distances after BAL encoding. We selected four pixels($n1$, $n2$, $n3$, $n4$) from each of the two categories($C1$ and $C2$) in both the BAL and one-hot label to compare the difference between the two label encoding. Intuitively, pixels closer to the boundary have larger variance for BAL. In the concluding part of this section, we will conduct a quantitative assessment of BAL's applicability within the model and explore strategies for satisfying the aforementioned constraints.

Finally, the BAL is specified as follows. We firstly obtain discrete Gaussian distribution vectors of $C$ dimensions. Then, the variance $\sigma_d$ and mean $\mu$ of the Gaussian distribution is determined by the pixel-to-edge distance $d$ within the same class of objects and the the ground truth respectively. Formally, let  $I \in \mathcal{R}^{H \times W \times 3}$  denotes the input images and its corresponding encoded label maps can be expressed as $Y=\left\{y^j_{i} \mid y^j_{i} \in \mathcal{R}^{W \times H\times C}\right\}_{j=0}^{j=C}$, where $j \in\{0, \ldots, C\}$ is the category of an object and $C$ is set to 50 as the maximum number of categories per image\cite{wang2021ainet}\cite{yang2020superpixel}, $i \in\{1, \ldots, N\}$ is the pixel index, and $N$ is the total number of pixels.  $y_i^j \in[0,1]$ is a continuous value instead of a binary value. 
Assuming a pixel is indexed as \(i=n\), and its ground truth is \(j=c\), the encoding label $y$ of the pixel can then be represented as follows:
\begin{equation}
y^c_{n}=\frac{1}{\sqrt{2 \pi}\sigma_d} exp{\left(-\frac{(x-c)^2}{2 \sigma^2_d}\right)}. 
\end{equation} 
The labels of the same category have identical means, ensuring that semantic information is not disrupted. And the relationship between mean values and categories will be discussed at the end of this section. The variance $\sigma_d$ is calculated based on the distance $d$ of each pixel $i$ to its boundaries. Inspired by the CSF model in the human visual system, we form relationships as follows.

\begin{equation}   
\sigma_d=\beta * e^{-d^{\alpha}},
\label{m_n}
\end{equation}
where $\beta$ and $\alpha$ are both hyperparameters, respectively associated with the maximum intra-class distance difference and the intra-class distance decay rate. In this paper, we set  $\beta$=1.2 and $\alpha$=0.5, respectively.
As shown in the Figure \ref{sima2boundary}, the three categories in Figure \ref{sima2boundary}(a) exhibit high color similarity and texture blurring, attributed to factors such as insufficient lighting. Nevertheless, the conventional one-hot labels treat pixels of different categories equally within objects, as depicted in Figure \ref{sima2boundary}(b). To provide a visual comparison highlighting the disparities between BAL and one-hot encoding,  Figure \ref{sima2boundary}(c) illustrates a heatmap depicting how variance changes with distance. The network experiences heightened loss near the object boundaries, urging the model to allocate greater attention to pixel relationships at the edges.

We adopt the cross-entropy loss to group pixels with similar attributes. Additionally, we use the $\ell_2$ norm to encourage spatial compactness within the superpixels. The full losses for network training comprise two components, i.e., cross-entropy (CE) and L2 reconstruction losses for the semantic label $l_s(p)$ and position vector $\mathbf{p}$, following \cite{yang2020superpixel}:

\begin{equation}
\mathcal{L}=\sum_p C E\left(l_s^{\prime}(p), l_s(p)\right)+\frac{m}{S}\left\|\mathbf{p}-\mathbf{p}^{\prime}\right\|_2,
\end{equation}
where S represents the superpixel sampling interval, and m is the weight balancing the two terms. And the $l_s^{\prime}(p)$ is obtained by reconstructing the semantic labels from the predicted association map Q.
\begin{figure*}
    \centering
    \begin{subfigure}[b]{0.3\textwidth}
        \includegraphics[width=\textwidth]{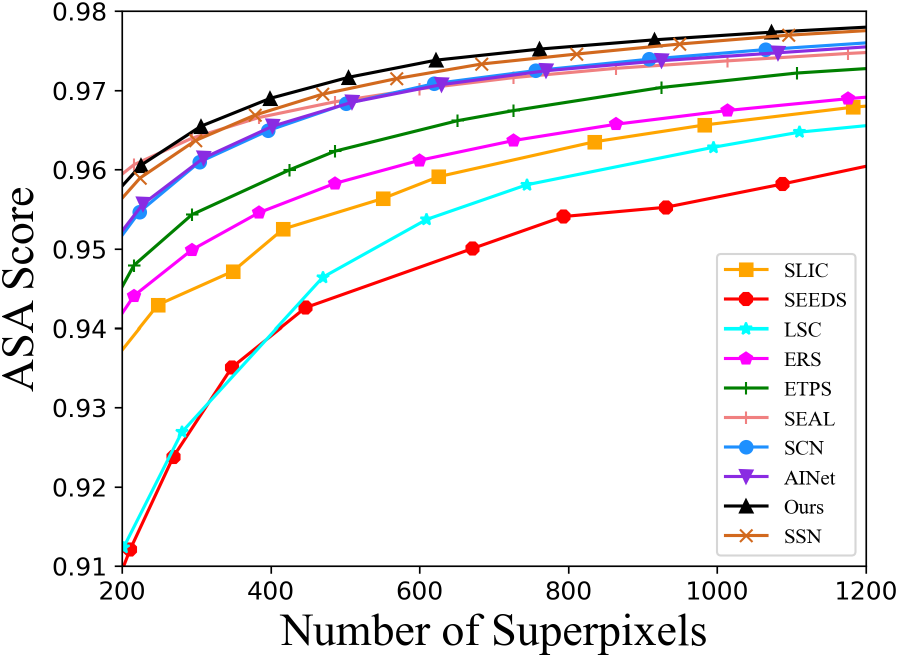}
        \caption{ASA on BSDS500}
        \label{fig:asa_bsds}
    \end{subfigure}
    ~ 
    \begin{subfigure}[b]{0.3\textwidth}
        \includegraphics[width=\textwidth]{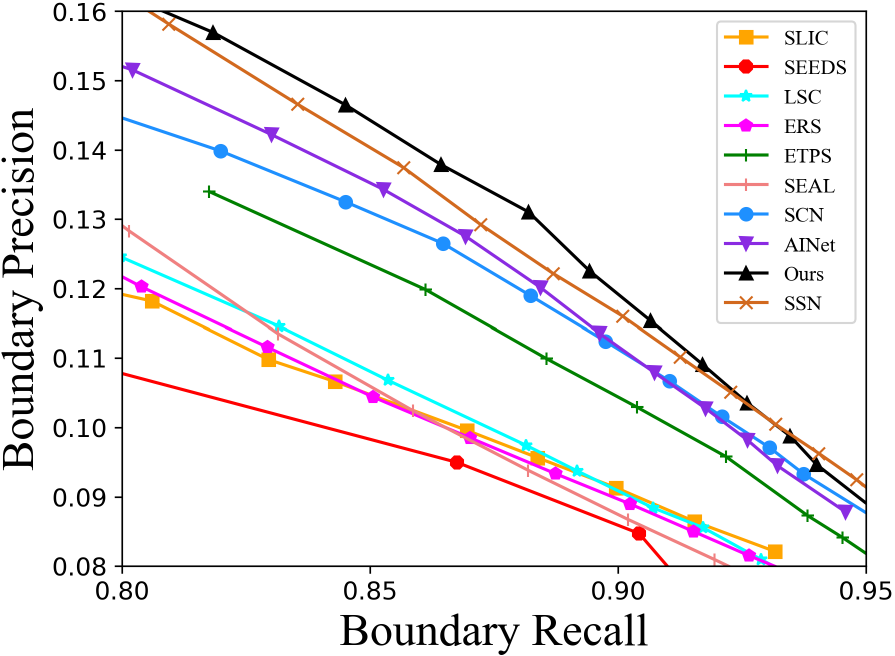}
        \caption{BR-BP on BSDS500}
        \label{fig:br_bsds}
    \end{subfigure}
    ~ 
    \begin{subfigure}[b]{0.3\textwidth}
        \includegraphics[width=\textwidth]{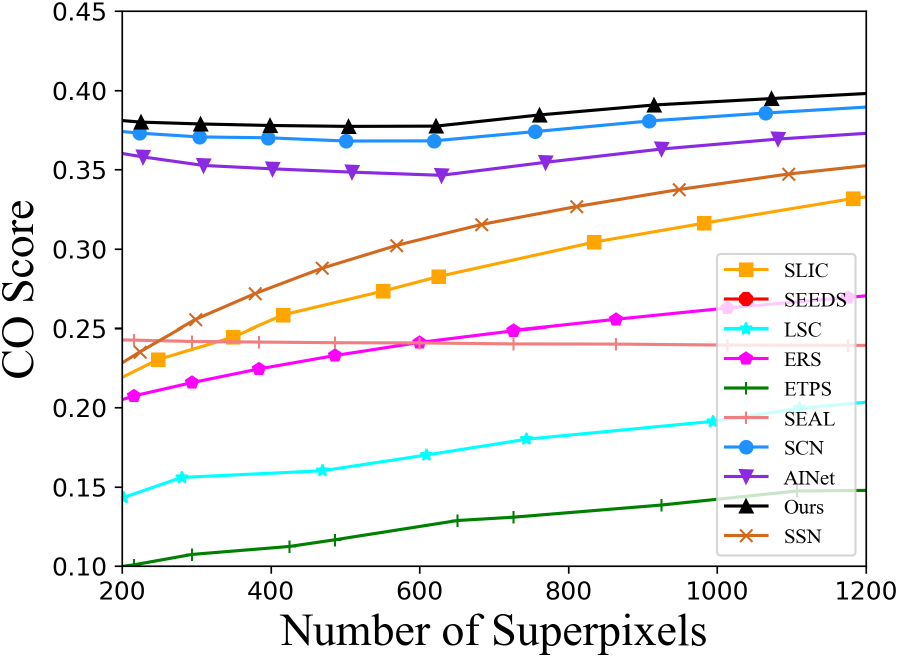}
        \caption{CO on BSDS500}
        \label{fig:co_bsds}
    \end{subfigure}
    \caption{Performance comparison on datasets BSDS500}
    \label{com_bsds}
\end{figure*}
\begin{figure*}
    \centering
    \begin{subfigure}[b]{0.3\textwidth}
        \includegraphics[width=\textwidth]{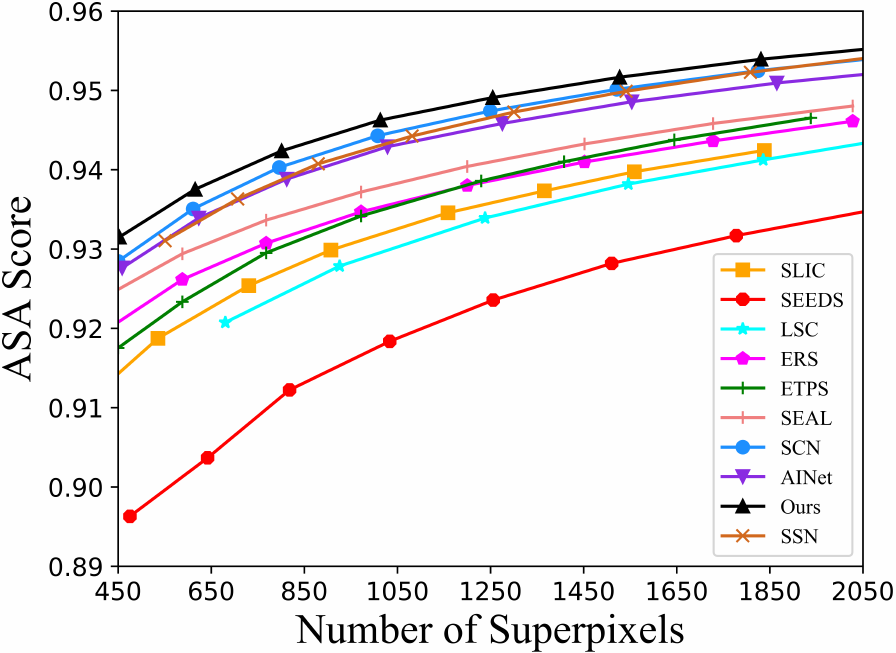}
        \caption{ASA on NYUv2}
        \label{fig:asa_nyu}
    \end{subfigure}
    ~ 
    \begin{subfigure}[b]{0.3\textwidth}
        \includegraphics[width=\textwidth]{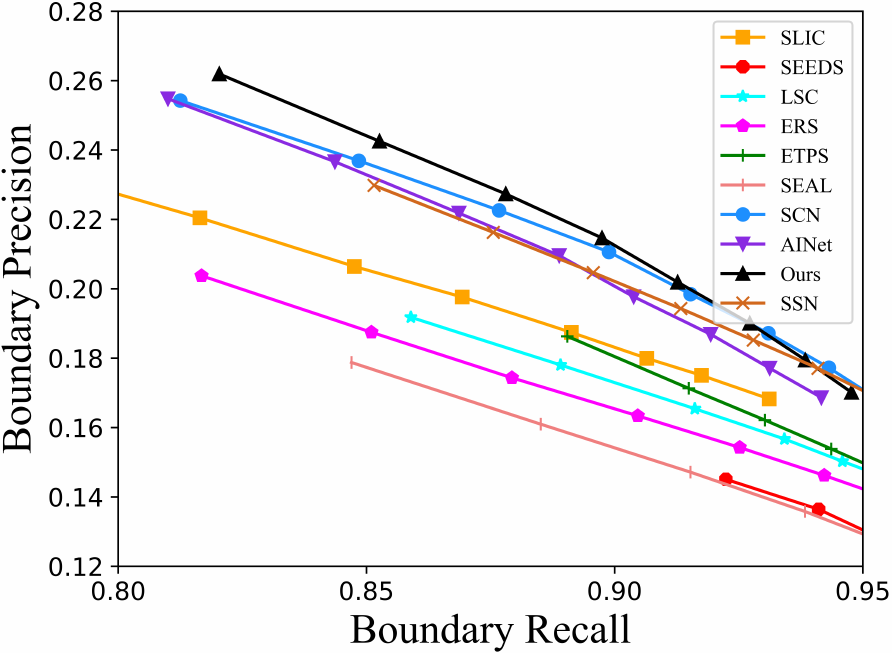}
        \caption{BR-BP on NYUv2}
        \label{fig:br_nyu}
    \end{subfigure}
    ~ 
    \begin{subfigure}[b]{0.3\textwidth}
        \includegraphics[width=\textwidth]{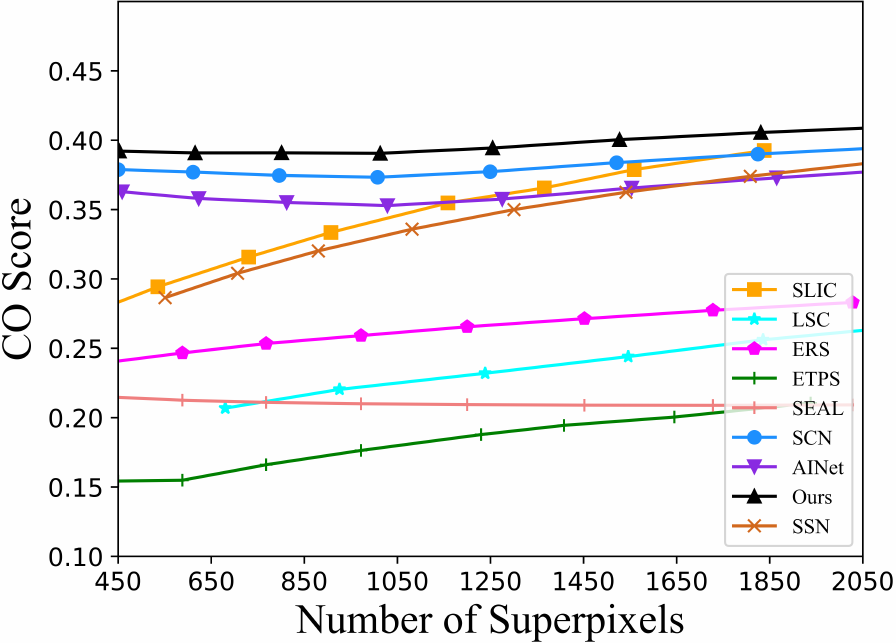}
        \caption{CO on NYUv2}
        \label{fig:co_nyu}
    \end{subfigure}
    \caption{Performance comparison on datasets NYUv2}
    \label{com_nyu}
\end{figure*}

\begin{figure*}[t]
    \centering
    \begin{subfigure}[b]{0.16\textwidth}
        \includegraphics[width=\textwidth]{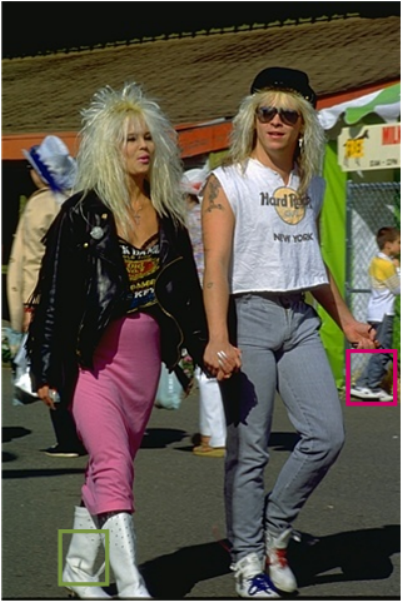}
        \vspace{1pt}
        \includegraphics[width=\textwidth]{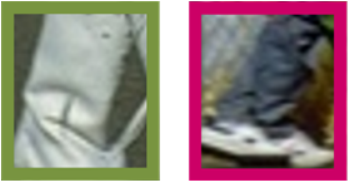}
        \vspace{2pt}
        \includegraphics[width=80pt,height=84pt]{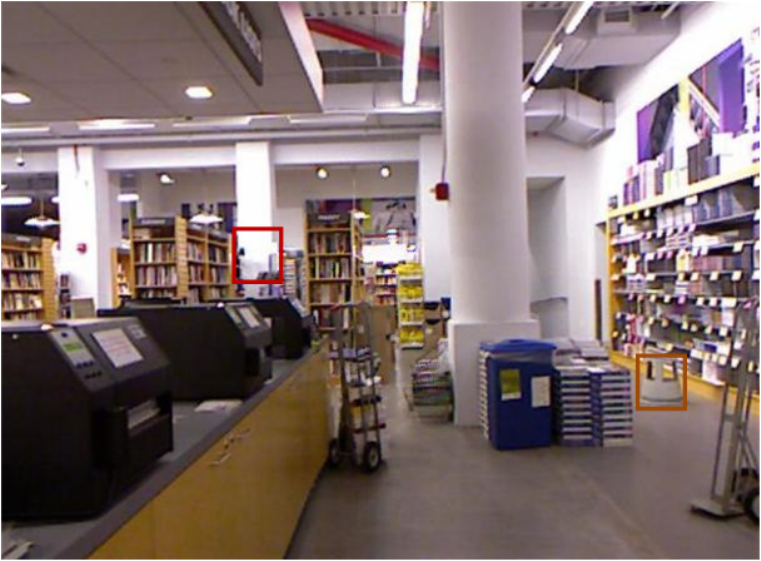}
        \vspace{1pt}
        \includegraphics[width=\textwidth]{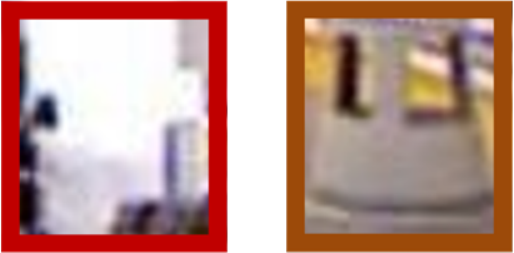}
        \caption{Input}
    \end{subfigure}
    \begin{subfigure}[b]{0.16\textwidth}
        \includegraphics[width=\textwidth]{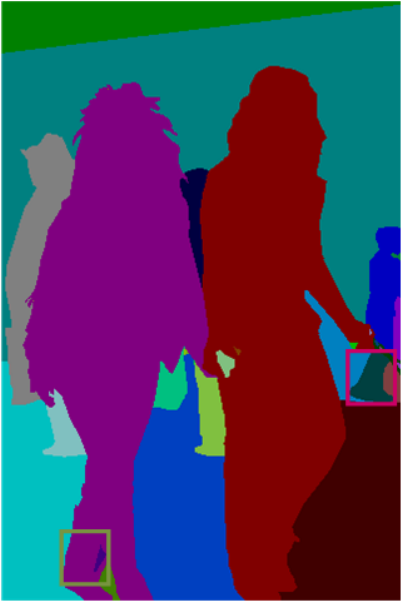}
        \vspace{1pt}
        \includegraphics[width=\textwidth]{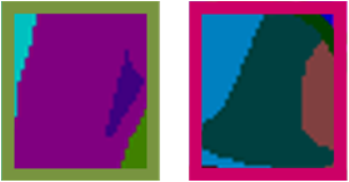}
        \vspace{2pt}
        \includegraphics[width=80pt,height=84pt]{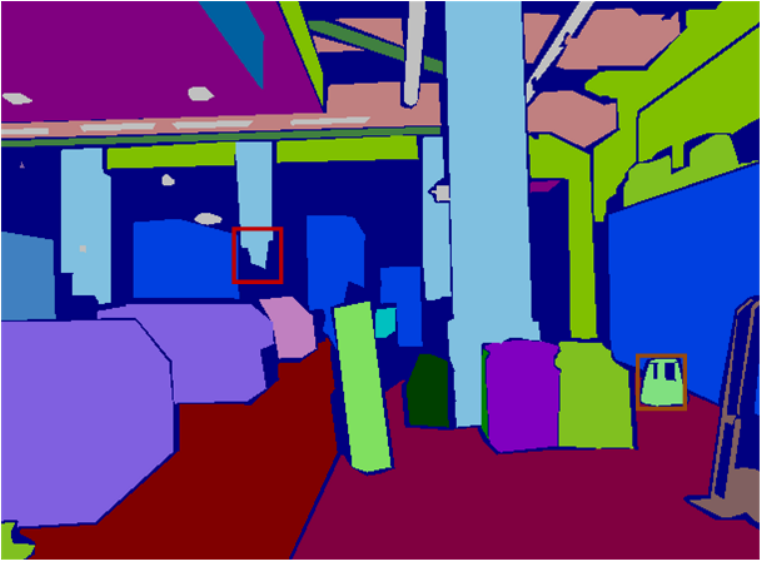}
        \vspace{1pt}
        \includegraphics[width=\textwidth]{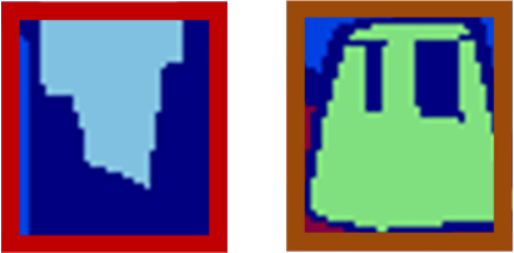}
        \caption{GT segments}
    \end{subfigure}
    \begin{subfigure}[b]{0.16\textwidth}
        \includegraphics[width=\textwidth]{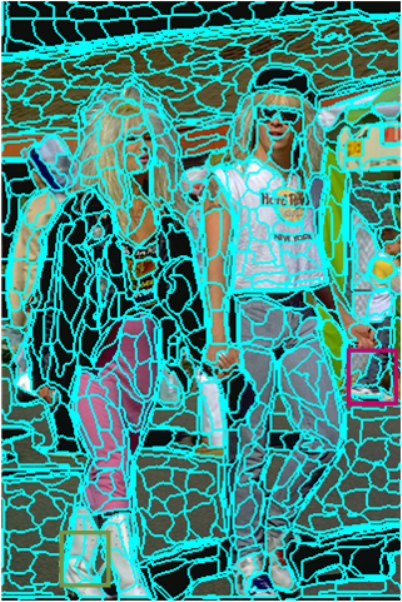}
        \vspace{1pt}
        \includegraphics[width=\textwidth]{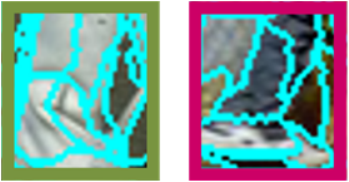}
        \vspace{2pt}
        \includegraphics[width=80pt,height=84pt]{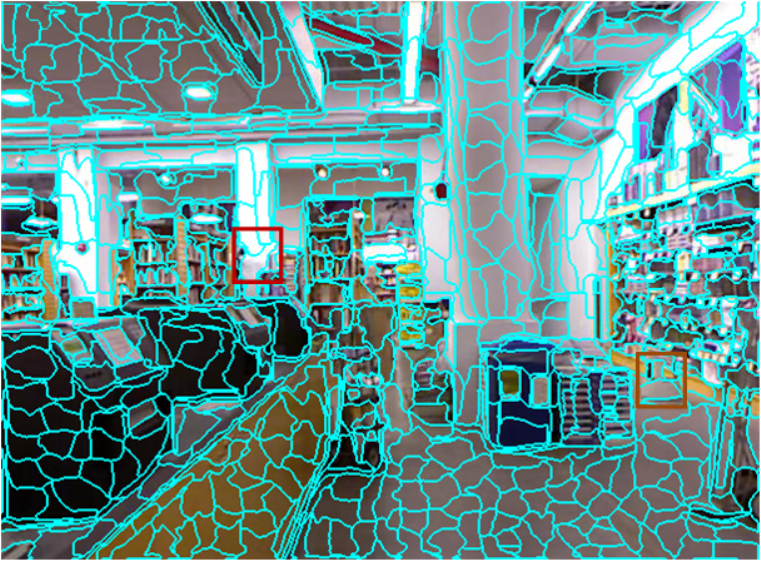}
        \vspace{1pt}
        \includegraphics[width=\textwidth]{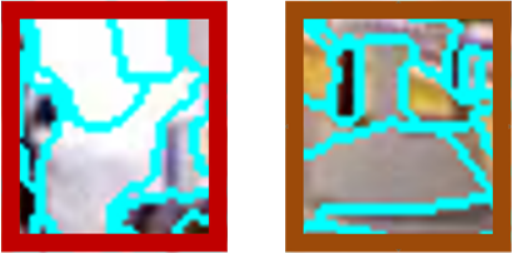}
        \caption{SSN}
    \end{subfigure}
    \begin{subfigure}[b]{0.16\textwidth}
        \includegraphics[width=\textwidth]{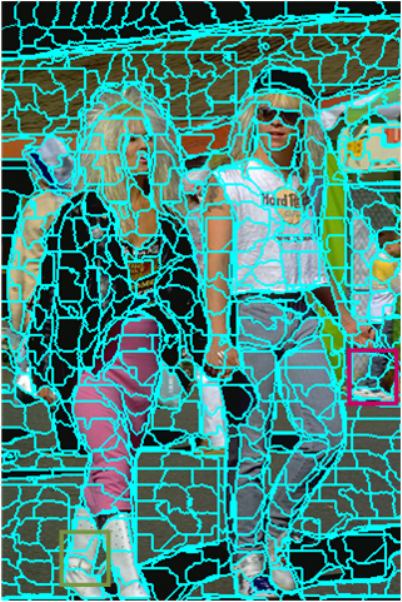}
        \vspace{1pt}
        \includegraphics[width=\textwidth]{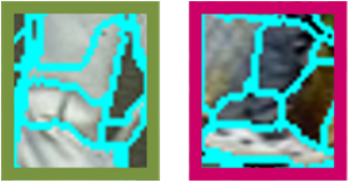}
        \vspace{2pt}
        \includegraphics[width=80pt,height=84pt]{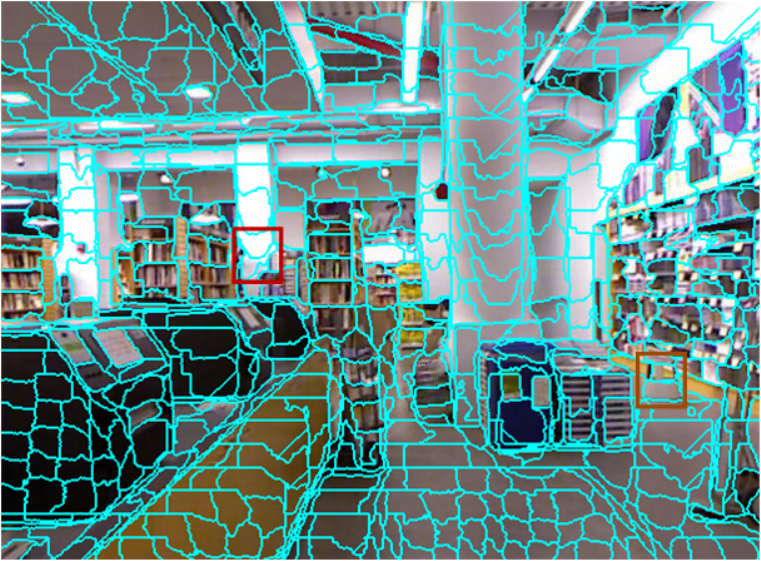}
        \vspace{1pt}
        \includegraphics[width=\textwidth]{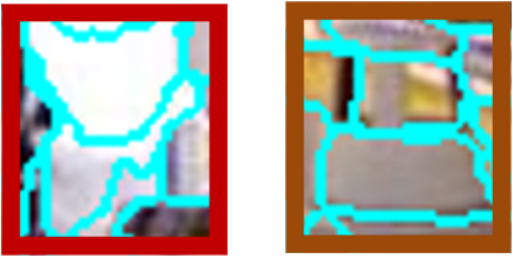}
        \caption{SCN}
    \end{subfigure}
    \begin{subfigure}[b]{0.16\textwidth}
        \includegraphics[width=\textwidth]{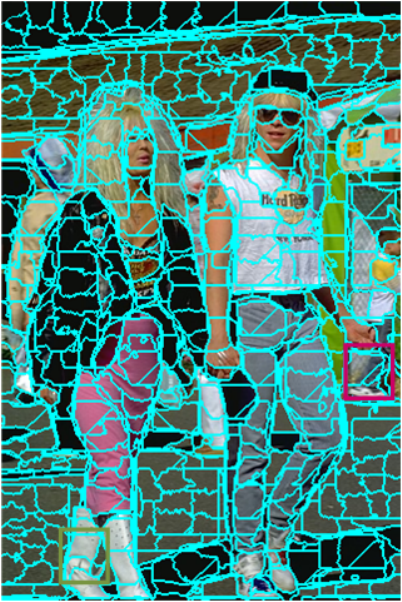}
        \vspace{1pt}
        \includegraphics[width=\textwidth]{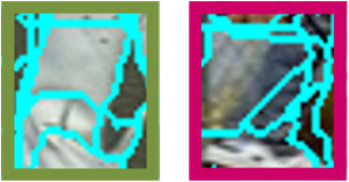}
        \vspace{2pt}
        \includegraphics[width=80pt,height=84pt]{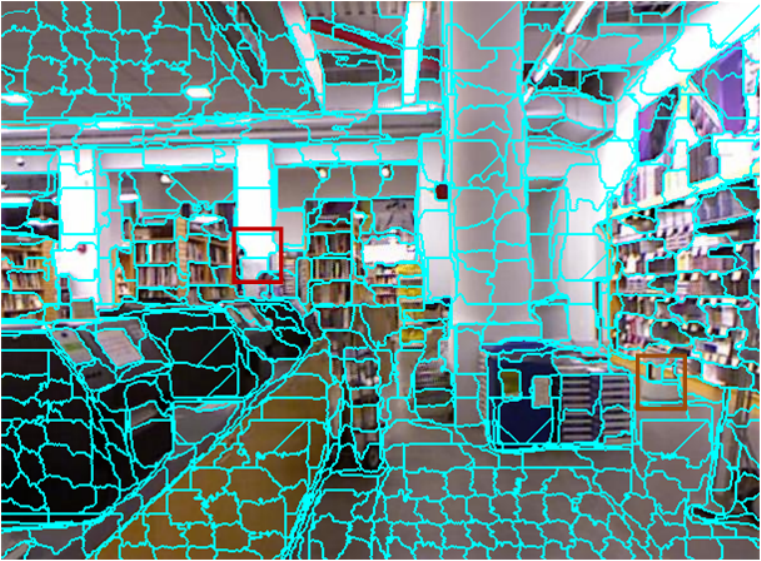}
        \vspace{1pt}
        \includegraphics[width=\textwidth]{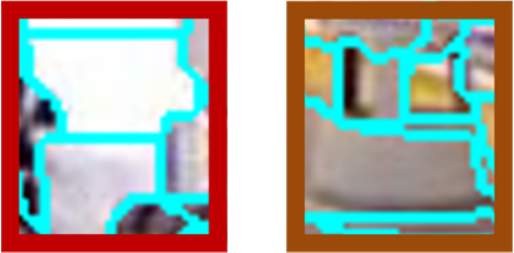}
        \caption{AINet}
    \end{subfigure}
    \begin{subfigure}[b]{0.16\textwidth}
        \includegraphics[width=\textwidth]{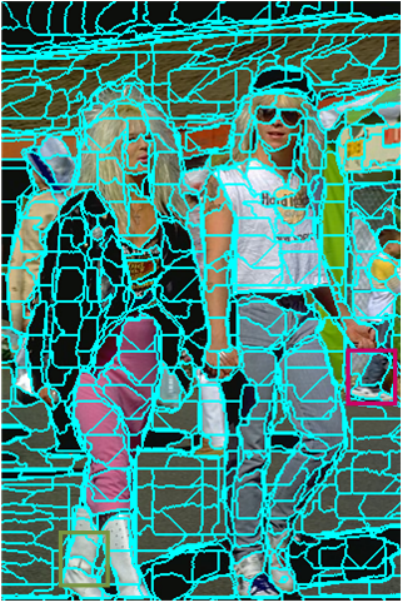}
        \vspace{1pt}
        \includegraphics[width=\textwidth]{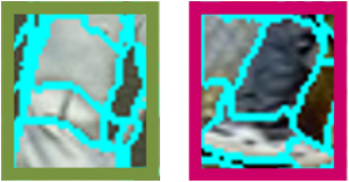}
        \vspace{2pt}
        \includegraphics[width=80pt,height=84pt]{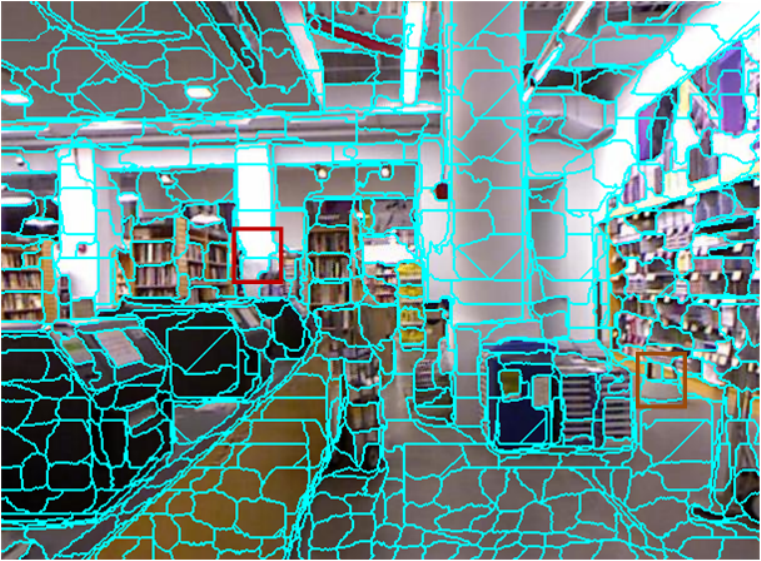}
        \vspace{1pt}
        \includegraphics[width=\textwidth]{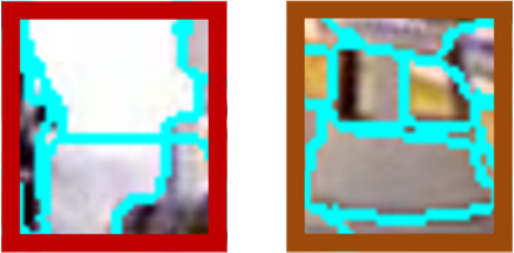}
        \caption{Ours}
    \end{subfigure}
    \caption{Qualitative results of four SOTA superpixel methods, SSN, SCN, AINet, and our method. The top row exhibits the results from BSDS500 dataset, while the bottom row shows the superpixels on NYUv2 dataset.}
    \label{spix_viz}
\end{figure*}

\begin{figure*}
    \centering
    \begin{subfigure}[b]{0.3\textwidth}
        \includegraphics[width=\textwidth]{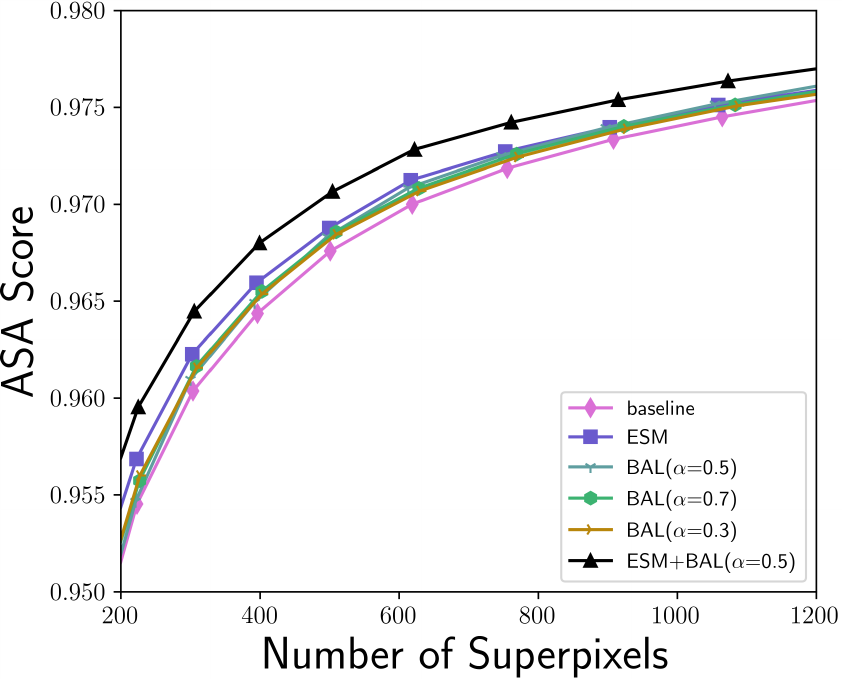}
        \caption{ASA on BSDS500}
    \end{subfigure}
    ~ 
    \begin{subfigure}[b]{0.3\textwidth}
        \includegraphics[width=\textwidth]{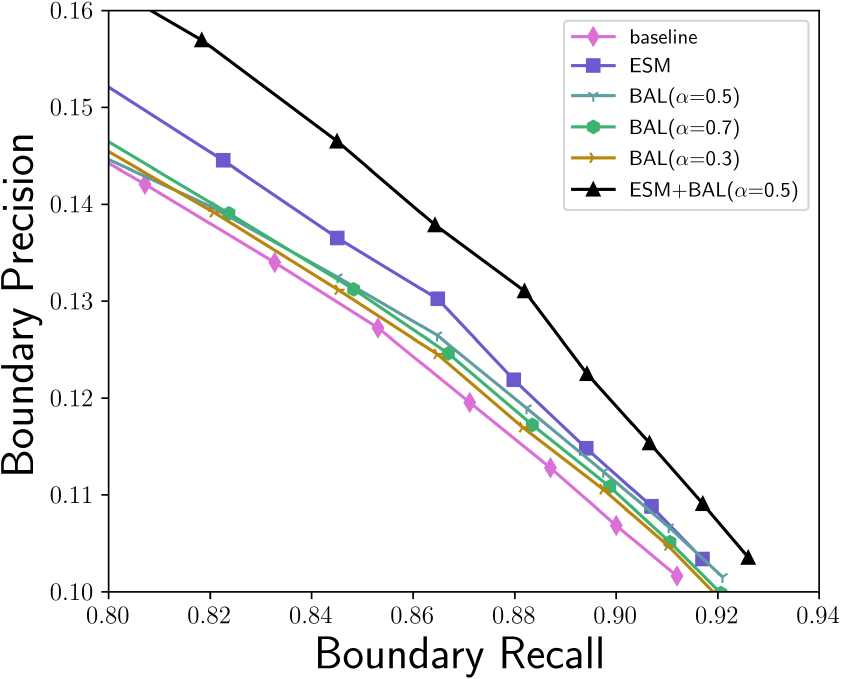}
        \caption{BR-BP on BSDS500}
    \end{subfigure}
    ~ 
    \begin{subfigure}[b]{0.3\textwidth}
        \includegraphics[width=\textwidth]{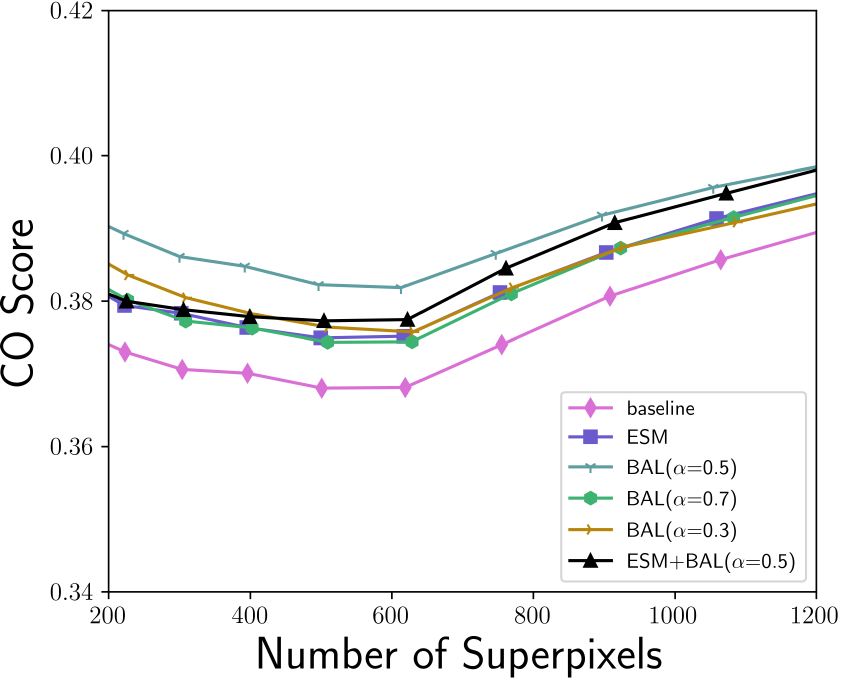}
        \caption{CO on BSDS500}
    \end{subfigure}
    \caption{Ablation study on BSDS500.}
    \label{Ablation}
\end{figure*}

\begin{figure*}[t]
    \centering
    \begin{subfigure}[b]{0.16\textwidth}
        \includegraphics[width=\textwidth]{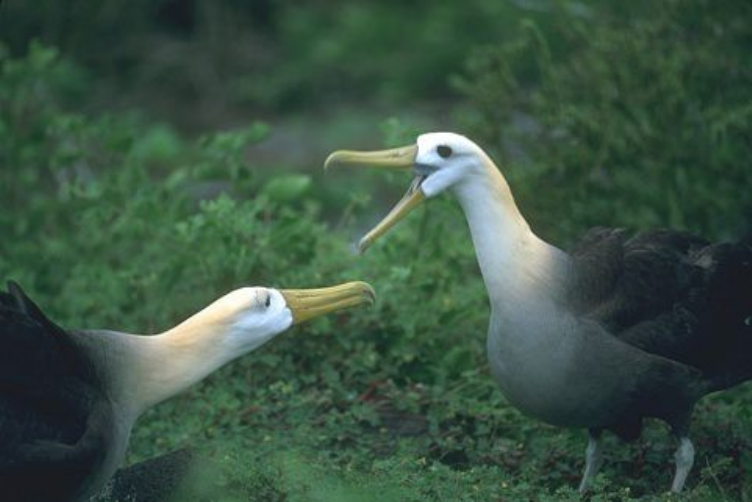}
        \includegraphics[width=\textwidth]{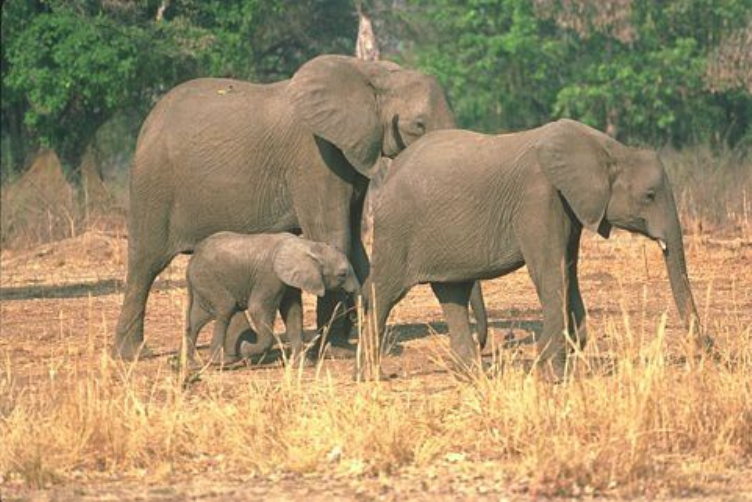}
        \includegraphics[width=\textwidth]{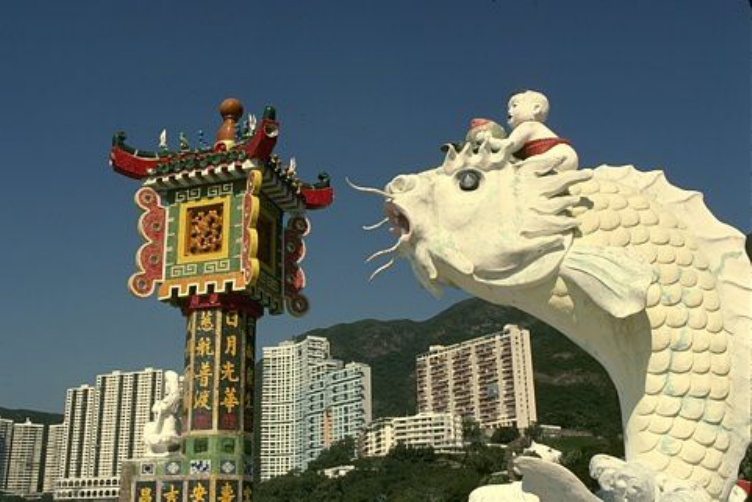}
        \caption{Images}
    \end{subfigure}
    \begin{subfigure}[b]{0.16\textwidth}
        \includegraphics[width=\textwidth]{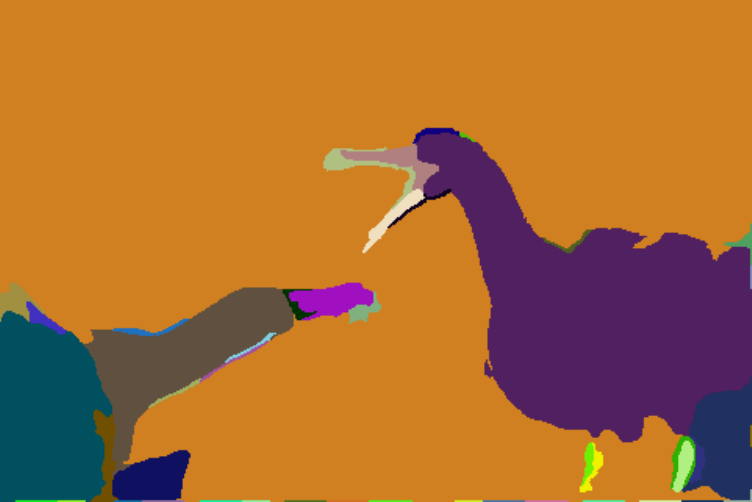}
        \includegraphics[width=\textwidth]{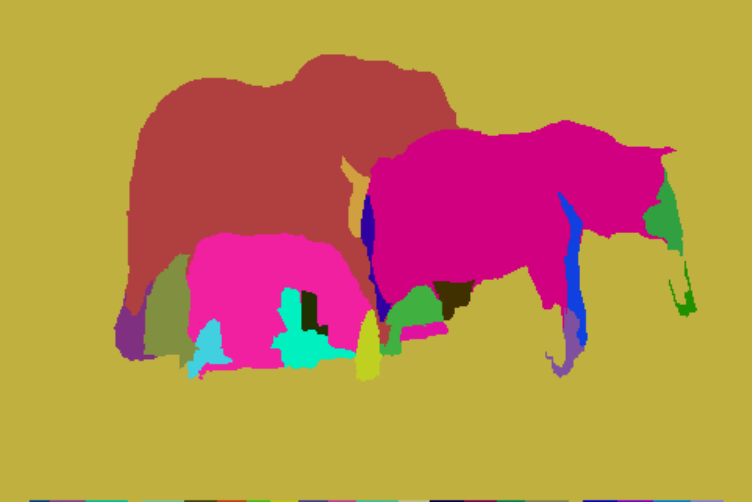}
        \includegraphics[width=\textwidth]{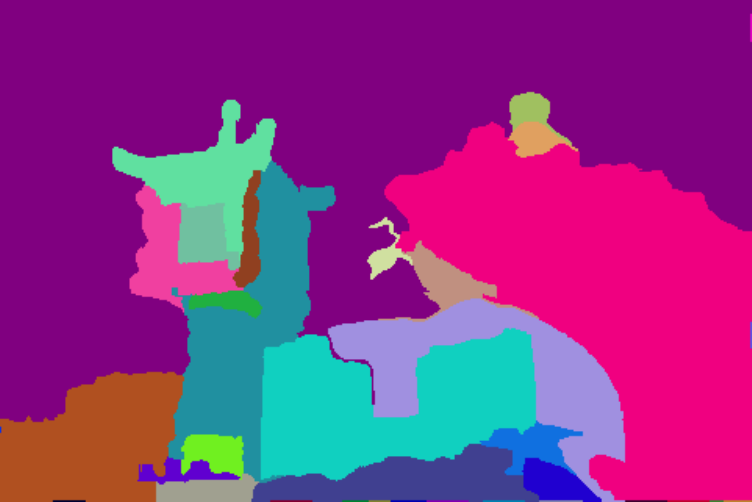}
        \caption{SSN\cite{jampani2018superpixel}}
    \end{subfigure}
    \begin{subfigure}[b]{0.16\textwidth}
        \includegraphics[width=\textwidth]{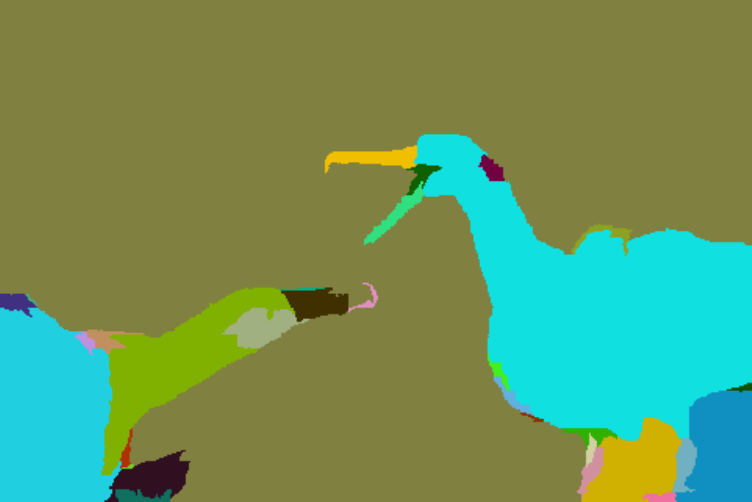}
        \includegraphics[width=\textwidth]{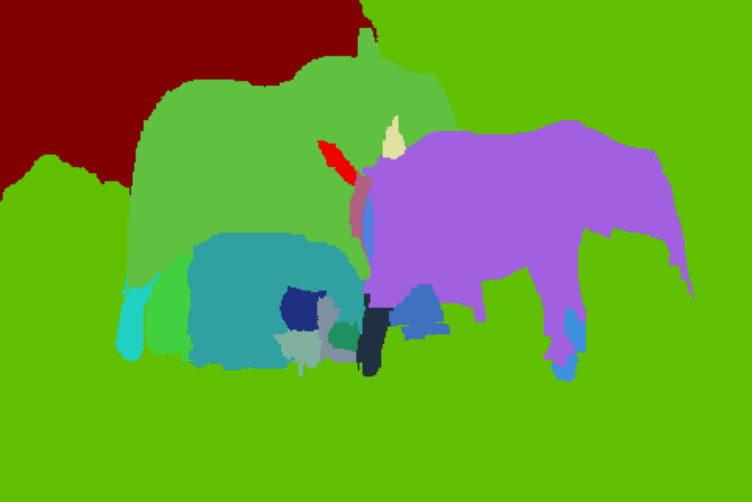}
        \includegraphics[width=\textwidth]{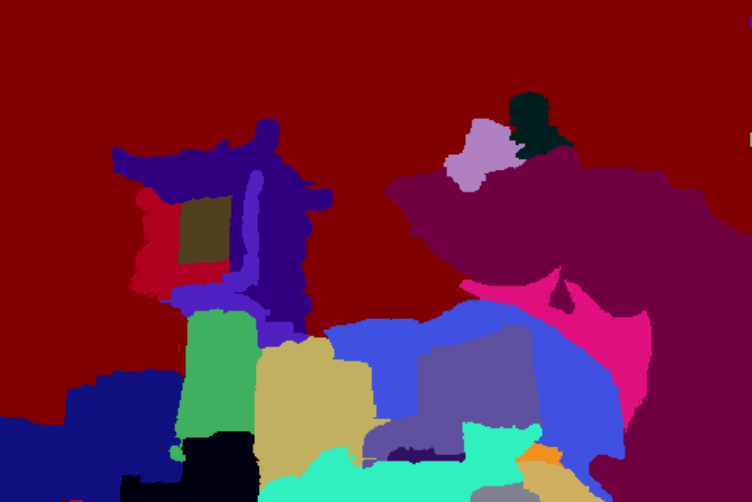}
        \caption{SCN\cite{yang2020superpixel} }
    \end{subfigure}
    \begin{subfigure}[b]{0.16\textwidth}
        \includegraphics[width=\textwidth]{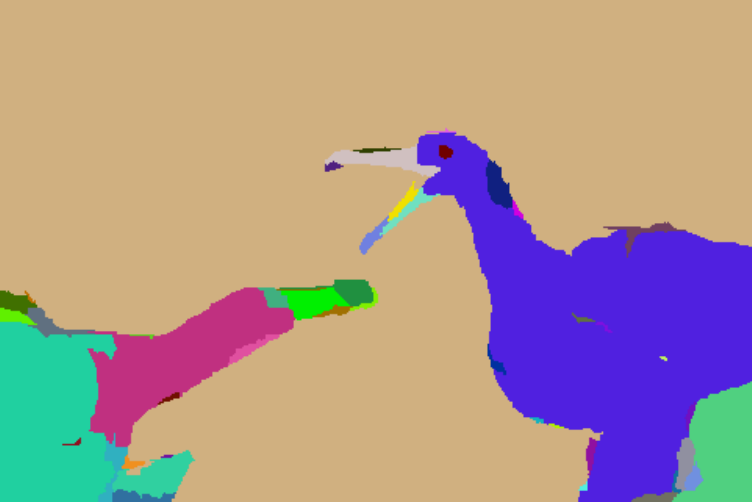}
        \includegraphics[width=\textwidth]{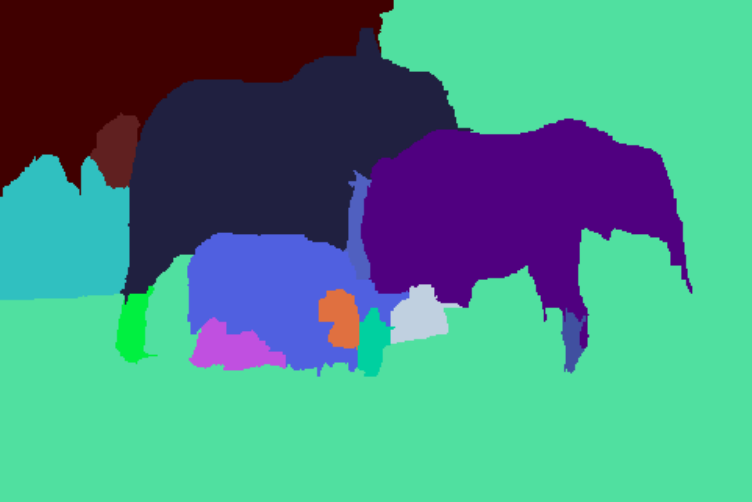}
        \includegraphics[width=\textwidth]{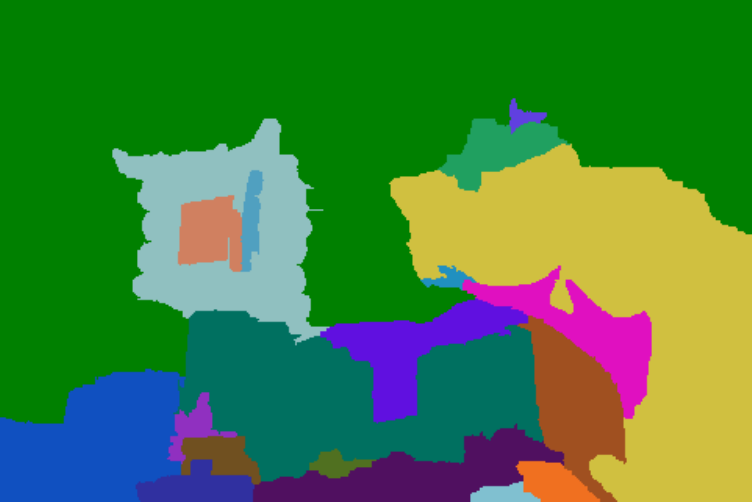}
        \caption{AINet\cite{wang2021ainet}}
    \end{subfigure}
    \begin{subfigure}[b]{0.16\textwidth}
        \includegraphics[width=\textwidth]{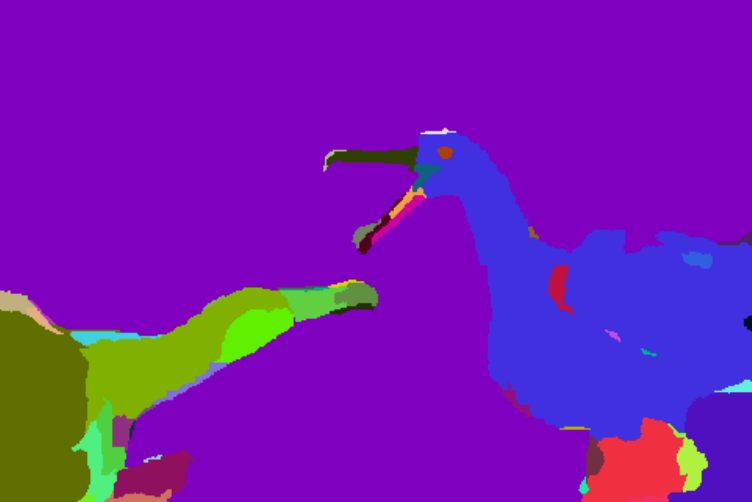}
        \includegraphics[width=\textwidth]{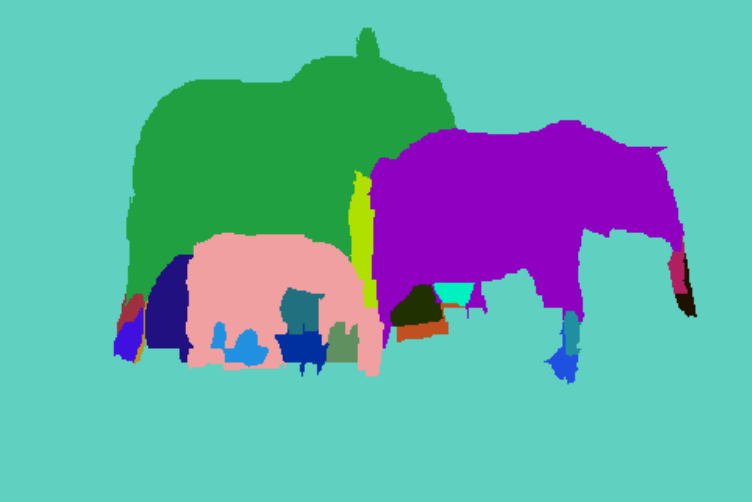}
        \includegraphics[width=\textwidth]{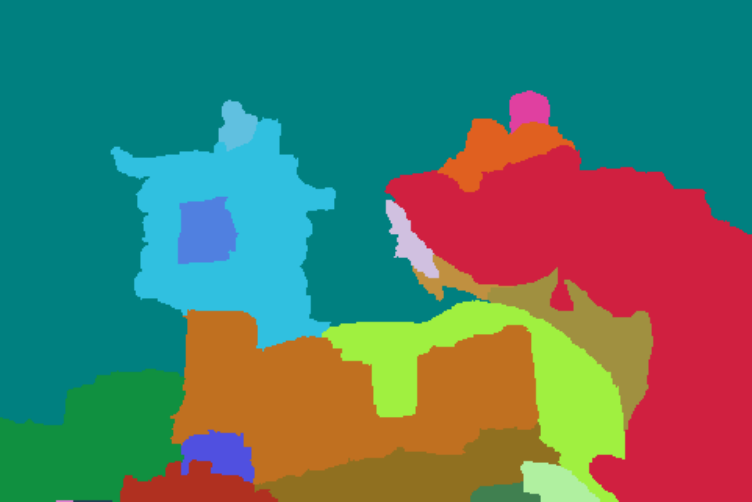}
        \caption{Ours}
    \end{subfigure}
    \begin{subfigure}[b]{0.16\textwidth}
        \includegraphics[width=\textwidth]{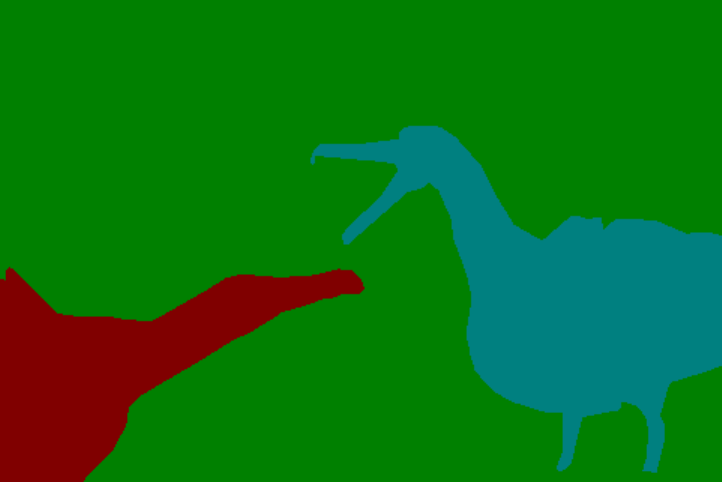}
        \includegraphics[width=\textwidth]{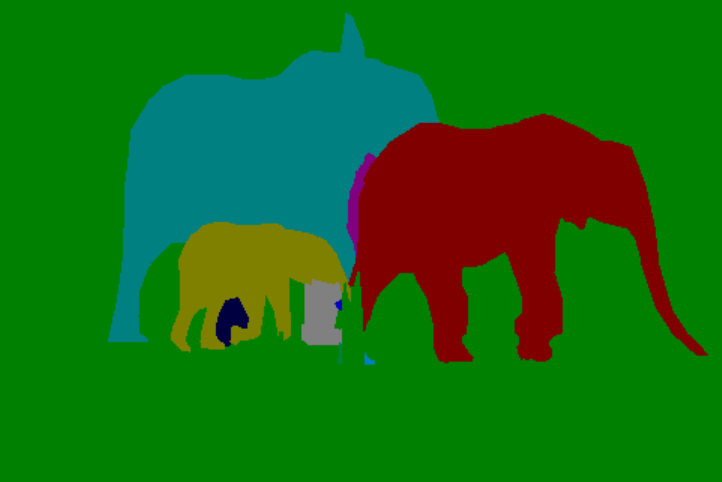}
        \includegraphics[width=\textwidth]{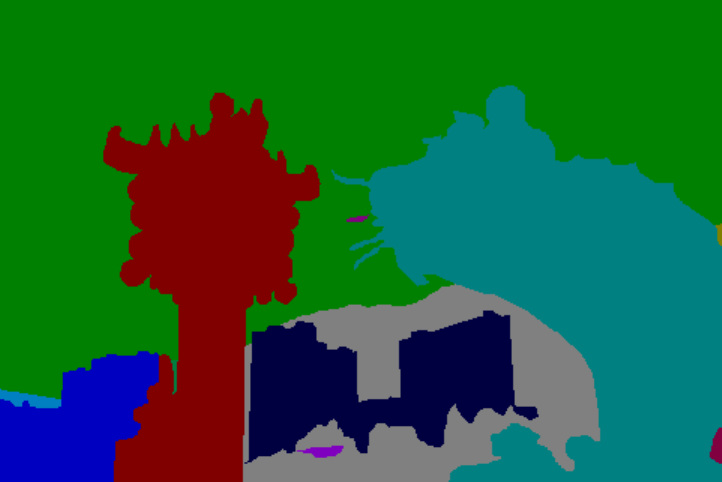}
        \caption{GT label}
    \end{subfigure}
    \caption{The generated proposals from DEL\cite{liu2018deep}using different superpixels.}
    \label{del_viz}
\end{figure*}
To visually illustrate the influence of varying variances and means on the model, we conduct a quantitative analysis of the relationships between different mean intervals($\Delta_\mu$) and variance intervals($\Delta_\sigma$) concerning the cross-entropy loss. Considering the size of the training data, the variance values are selected to be approximately within the range of 0.3 to 1.2. As depicted in Figure \ref{sigmaandmean}, Figure\ref{sigmaandmean}(a) reveals that when the mean interval exceeds 10, the loss no longer varies. Consequently, we set the $\Delta_\sigma$ between different categories to be equal to 10. ensuring that the loss between any two categories is the same, consistent with one-hot encoding label. Figure\ref{sigmaandmean}(b) clarifies the relationship between varying $\Delta_\sigma$ and the loss. It becomes evident that the fluctuations in loss intervals, ranging from 0.05 to 0.8, closely approximate a linear progression. As a result, we set $\Delta_\mu = 10$ , regardless of the varying variances' impact on the model, as mentioned earlier.

\section{Experiment}
\label{sec:typestyle}
\subsection{Datasets and Implementation details}
To evaluate our approach's efficiency, we perform a series of experiments on the BSDS500\cite{arbelaez2010contour} and NYUv2\cite{silberman2012indoor} benchmark datasets. The BSDS500 dataset contains 200 training, 100 validation, and 200 testing images.Considering each image annotation as an independent sample, there are 1,087 training samples, 546 validation samples, and 1,063 testing samples in BSDS500, as adopted from \cite{yang2020superpixel} \cite{wang2021ainet} \cite{jampani2018superpixel}\cite{tu2018learning}. The 1087 training samples were used to train the biomimetic network model. The NYUv2 dataset, originally applied to indoor scene understanding tasks, comprises 1,449 images labeled with object instances. \cite{stutz2018superpixels} excluded data from unlabeled boundary regions and selected a subset of 400 test images for superpixel evaluation, with each image measuring 608$\times$448 pixels.

To assess data generalization, we adopt the approach of AINet\cite{wang2021ainet} and SCN\cite{yang2020superpixel}. Initially, we train our model on the BSDS500 dataset. Subsequently, this model is applied to the NYUv2 dataset without any fine-tuning. Additionally, we execute SLIC, SEEDS, LSC, ERS, and ETPS using the codes and optimal parameters provided in \cite{stutz2018superpixels}. Consequently, the results of partial methods presented in this paper vary from those in SCN\cite{yang2020superpixel} and AINet\cite{wang2021ainet} Furthermore, we run SEAL, SCN\cite{yang2020superpixel}, and AINet\cite{wang2021ainet} using the original weights and code repositories provided  by their respective authors.

During the training process, input images are randomly cropped to dimensions of 208$\times$208 for the BSDS500 dataset. The initial learning rate is set at 8e-5 and is reduced by a factor of 0.5 after 8,000 iterations. The model is optimized utilizing the Adam optimization method and is trained over 200 epochs with a batch size of 8. Experiments are conducted within the PyTorch framework on a workstation equipped with an Intel Core i7 CPU and an RTX 2080 GPU. A sampling interval of 16 is in alignment with existing literature. For testing, a consistent strategy is employed to generate superpixel grids, as referenced in \cite{yang2020superpixel}.

\subsection{Comparison with the state-of-the-arts}
We evaluate the performance of superpixel methods using three metrics, including achievable segmentation accuracy(ASA), boundary recall and precision(BR-BP) and compactness (CO).
A comprehensive performance comparison on BSDS500 and NYUv2 test sets is illustrated in Figures \ref{com_bsds} and \ref{com_nyu}. Given that traditional superpixel segmentation algorithms are markedly outperformed by deep neural network-based ones, we include only the representative algorithms, such as, SLIC, LSC, SEEDS, etc for comparison. Observing from Figure\ref{com_bsds}(a)-(c), the proposed biomimetic network based on biological visual mechanisms, achieves the best results on the BSDS500 dataset in terms of ASA, BR-BP and CO. This suggests that our model can accurately extract inter-object hierarchical relationships, regardless of the number of segmented superpixels. Figures\ref{com_nyu}(a)-(c) depict the test results on the NYU2 dataset without fine-tuning the model. Our model maintains its commendable performance, signifying its robustness and generalizability. The Figure\ref{spix_viz} shows the qualitative results of four state-of-the-art methods on dataset BSDS500 and NYUv2, comparing to the competing methods, the boundaries of our results are more accurate and clearer, which intuitively shows the superiority of our method.
\begin{figure}
	\includegraphics[width=0.85\linewidth]{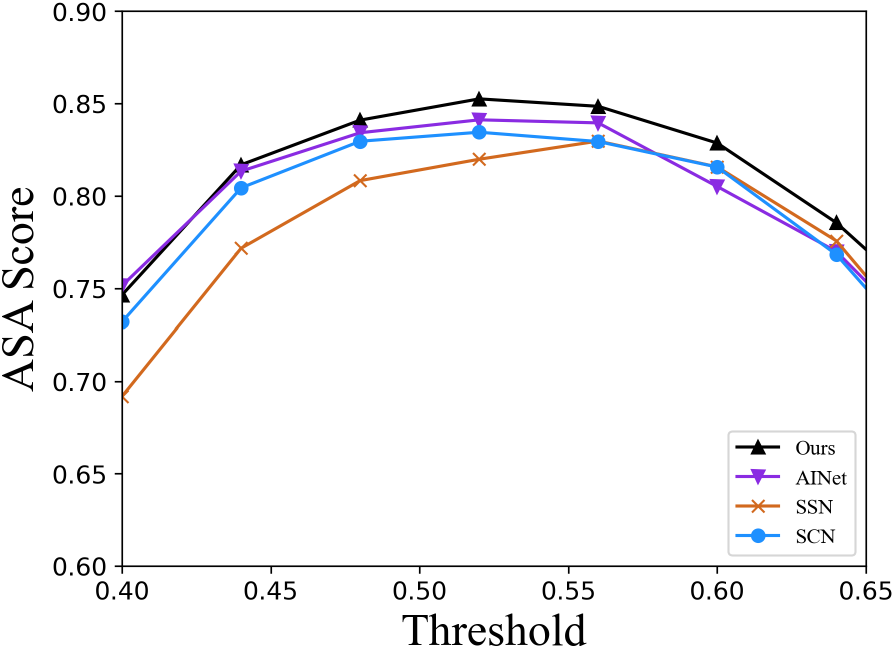}
	\centering
	\caption{The ASA scores of four state-of-the-art methods on object proposal generation.}
	\label{fig:asa_del}
\end{figure}
\subsection{Ablation Study}
We conduct comprehensive experiments to evaluate the effectiveness of each component in the proposed methods. The  performances of all ablation experiments on BSDS500 are shown in Figure \ref{Ablation}, where ESM stands for the Enhanced Screening Module of the biomimetic Network, and the BAL means the Boundary-Aware Label encoding strategy.
For the BAL method, we conduct experiments with various values of $\alpha$ in Eq.\ref{m_n}, namely, $\alpha = 0.3$, $\alpha = 0.5$ and $\alpha = 0.7$, in order to evaluate how different decay rates affect the experimental outcomes.
The Encoder-Decoder network serves as our baseline. It's evident that incorporating the proposed ESM and BAL inspired by biological vision mechanisms to the baseline method delivers remarkable enhancements. The proposed ESM appears to be the most contributive towards these evaluation metrics. BAL shows a slightly lower improvement in ASA and BR-PR but performs better on CO, especially when the number of superpixels is small. This could be attributed to the fact that when there is a higher number of superpixels, the model needs to pay increased attention to boundary information, resulting in reduced compactness. The combined effect of ESM and BAL has led to an improvement in segmentation precision and recall, albeit with a slight compromise in compactness. This underscores the potential significance of designing network algorithms from a human visual characteristics standpoint, especially when considering that image labels are annotated by different experts.

\subsection{Application on Object Proposal Generation}
Superpixels provide semantic boundaries for multiple regions, preserving essential image attributes. Therefore, superpixels can generate object proposals for downstream tasks, leading to reduced annotation expenses. For instance,  Liu et al. proposed the DEL\cite{liu2018deep} to compute the similarity between superpixels and merges them into object proposals based on various thresholds. In this section, we utilize the DEL\cite{liu2018deep} framework to visualize the generated proposals of superpixels from several different algorithms, such as SCN, SSN, AINet and our methods. We employ ASA score to measure how well the produced object proposals cover the ground-truth labels.

Figure \ref{del_viz} illustrates the results of four distinct superpixel segmentation methods within the DEL\cite{liu2018deep} framework, all employing the same threshold for superpixel merging. From Figure \ref{del_viz}, it can be observed that our method preserves more fine-grained details under the same granularity, such as the goose eyes in the first row of the Figure \ref{del_viz}. And it avoids meaningless over-segmentation, such as the tower in the third row. As depicted in Figure \ref{fig:asa_del}, the ASA score demonstrates that our approach outperforms other methods with different thresholds. In conclusion, our approach could generate more satisfactory object proposals comparing to the competing methods, which validates the effectiveness of our proposed method. 

\section{Conclusion}
Deep neural network-based algorithms still retain challenges when relying on limited training data. In this paper, we propose a novel biomimetic network architecture for superpixel segmentation, drawing inspiration from the human neural structure and visual mechanisms. Two main contributions are the Enhanced Screening Module(ESM) and the Boundary-Aware Label(BAL). The ESM mirrors the multilevel synchronization hypothesis of the visual cortex, merging both low and high-level feature maps. This integration and calibration of feature results in more accurate pixel-to-region relationships. Meanwhile, the proposed BAL is inspired by the band-pass frequency sensitivity observed in the human visual cortex. It amplifies the network's response to boundaries, enabling the production of precise superpixels that are semantically coherent. Our work underscores the potential of combining human neurobiological knowledge with computer vision, presenting a pioneering step towards more human-aligned superpixel segmentation techniques.



\bibliographystyle{cas-model2-names}

\bibliography{manuscript}



\end{document}